\documentclass[11pt]{article}

\usepackage[preprint]{acl}

\usepackage{hyperref}
\usepackage{graphicx}
\usepackage{tcolorbox}
\usepackage{adjustbox}
\usepackage{graphicx}
\usepackage{ragged2e}
\usepackage{makecell}
 
\usepackage{enumitem}
\usepackage{float}
\usepackage{xurl}

\usepackage{booktabs}
\usepackage{siunitx}
\usepackage{subcaption} 
\usepackage{xcolor}
\usepackage{booktabs}
\usepackage{tabularx}
\usepackage{xcolor}
\usepackage[table]{xcolor}
\usepackage{xspace}

\usepackage{times}
\usepackage{latexsym}

\usepackage[T1]{fontenc}

\usepackage[utf8]{inputenc}

\usepackage{microtype}

\usepackage{inconsolata}

\usepackage{graphicx}

\title{Scaling Test-Time Compute to Achieve IOI Gold Medal\\ with Open-Weight Models}

\author{Mehrzad Samadi, Aleksander Ficek, Sean Narenthiran,  Siddhartha Jain  \\ \textbf{Wasi Uddin Ahmad, Somshubra Majumdar, Vahid Noroozi, Boris Ginsburg} \\
NVIDIA\\
\texttt{\{msamadi,aficek,snarenthiran,siddjain}\\
\texttt{wasiuddina,smajumdar,vnoroozi,bginsburg\}@nvidia.com}
}

\newcommand{\gencluster}{\textsc{GenCluster}\xspace}

\begin{document}
\maketitle

\begin{abstract}
Competitive programming has become a rigorous benchmark for evaluating the reasoning and problem-solving capabilities of large language models (LLMs). The International Olympiad in Informatics (IOI) stands out as one of the most prestigious annual competitions in competitive programming and has become a key benchmark for comparing human and AI-level programming ability. While several proprietary models have been claimed to achieve gold medal-level performance at the IOI, often with undisclosed methods, achieving comparable results with open-weight models remains a significant challenge. In this paper, we present \gencluster, a scalable and reproducible test-time compute framework that attains IOI gold-level performance using open-weight models. It combines large-scale generation, behavioral clustering, ranking, and a round-robin submission strategy to efficiently explore diverse solution spaces under limited validation budgets. Our experiments show that the performance of our proposed approach scales consistently with available compute, narrowing the gap between open and closed systems. Notably, we will show that \gencluster can achieve a gold medal at IOI 2025 for the first time with an open-weight model \texttt{gpt-oss-120b}, setting a new benchmark for transparent and reproducible evaluation of reasoning in LLMs. 
\end{abstract}

\section{Introduction}
\begin{figure*}[tb]
    \centering    
    \includegraphics[width=1.0\textwidth]{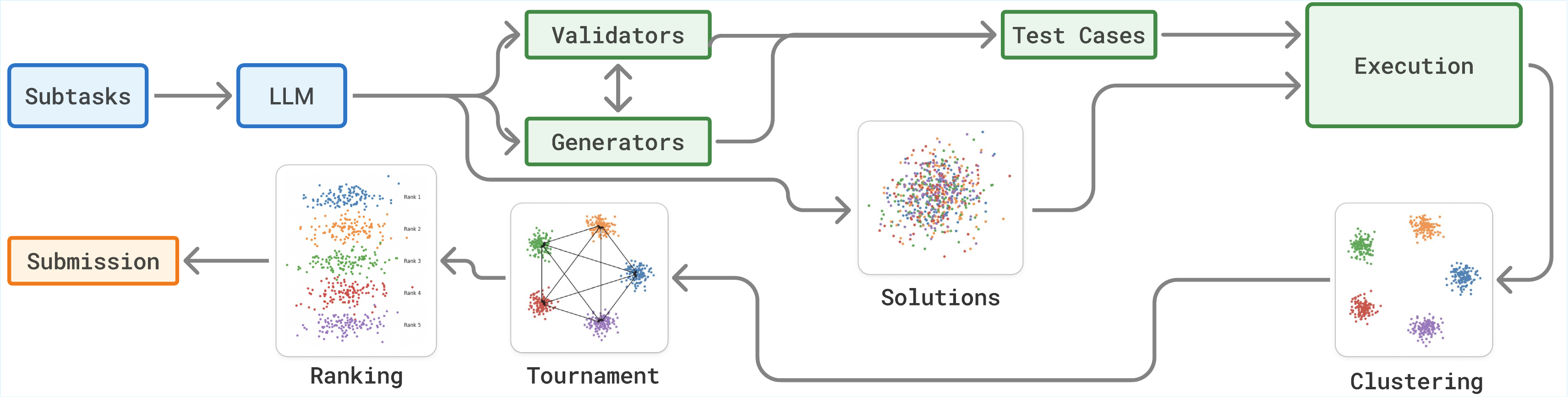}
    \caption{The overall pipeline of \gencluster for a  single subtask, a process to be repeated for every subtask in the IOI benchmark.}
    \label{fig:genclusteroverview}
\end{figure*}

As large language models (LLMs) have advanced in solving coding problems, traditional benchmarks such as HumanEval~\citep{chen2021codex_humaneval} and MBPP~\citep{austin2021mbpp} have reached saturation. This has motivated a shift toward more challenging benchmarks like LiveCodeBench~\citep{jain2025livecodebench} and Codeforces~\citep{openr1_codeforces}, which feature significantly more complex competitive programming problems. Furthermore, considering the challenging aspects of understanding and solving the competitive programming questions, they have been shown to be effective for both training and evaluating the reasoning capabilities of LLMs~\citep{deepseek_r1}. Recent progress on these benchmarks has been driven by both improved training strategies and test-time compute methods. For example, AlphaCode2~\citep{Leblond2023AlphaCode2} achieved performance exceeding that of 85\% of Codeforces participants by generating over one million candidate programs per problem and then applying a pipeline of filtering, clustering, and ranking to select 10 solutions for final submissions.

Analogous to the International Mathematical Olympiad (IMO)\footnote{https://www.imo-official.org/}, the International Olympiad in Informatics (IOI)\footnote{https://ioinformatics.org/} and International Collegiate Programming Contest (ICPC)\footnote{https://icpc.global/} are widely regarded as the pinnacle of algorithmic programming competitions. These competitions serve as an important milestone in assessing AI competency on competitive programming and general reasoning capabilities as a whole. OpenAI reported achieving a gold medal at IOI 2024 with their \textit{o1-ioi} and \textit{o3} models~\citep{openai2025competitiveprogramminglargereasoning}, using specialized test-time compute approaches. \textit{o1-ioi} is a dedicated version of their \textit{o1} model fine-tuned for competitive programming and leveraging external tools. Additionally, OpenAI claimed a gold medal and a 6th-place human-equivalent ranking at IOI 2025 with their latest models \citep{thedecoder_openai_ioi2025}, though the details of these systems have not yet been fully disclosed. Recently, both OpenAI and Google DeepMind~\citep{LinCheng2025GeminiICPC} have reported achieving gold-level performance at ICPC 2025 using proprietary methods. 

Despite their impressive results, the techniques and models used to achieve gold-level performance at such competitions are rarely disclosed. In contrast, open-weight models such as \texttt{DeepSeek-R1-0528}~\citep{deepseek_r1} and \texttt{Qwen3-235B-A22B}~\citep{qwen3} have achieved increasingly competitive results on LiveCodeBench and Codeforces, but still lag behind proprietary models. In this paper, we propose \gencluster, a scalable test-time compute approach that substantially improves the performance of open-weight LLMs on IOI problems. Our approach first generates a large pool of candidate solutions for each problem, which are then refined through a pipeline of filtering, behavioral clustering, and ranking with tournament. Finally, a round-robin submission strategy selects the best candidates while adhering to the same submission constraints imposed on human contestants at IOI (Figure \ref{fig:genclusteroverview}). In our experiments, we demonstrate that \gencluster enables open-weight model \texttt{gpt-oss-120b}~\citep{openai2025gptoss120bgptoss20bmodel} to achieve gold-level performance at IOI 2025. 

In our experiments, we evaluated and analyzed some of the best open-weight models publicly available including \texttt{DeepSeek-R1-0528} and \texttt{Qwen3-235B-A22B}, and we show that \texttt{gpt-oss-120b} is significantly superior on IOI problems, providing better scalability in terms of the number of generations. Furthermore, we demonstrate that \gencluster improves scores with increased compute and larger generation budgets which is an indicator of scalability of our proposed test-time compute approach.

In summary, our contributions are as follows:
\begin{enumerate}
    \item We propose \gencluster, a scalable test-time compute approach to improve the competitive programming performance of LLMs by generating many solutions in parallel and then using behavioral clustering and tournaments to select the best candidates.
    \item We demonstrate, for the first time, that \textbf{gold-level performance} at IOI 2025 can be achieved using only \textbf{open-weight models} combined with a transparent and reproducible \textbf{test-time compute} strategy. We plan to release our code publicly. \footnote{\url{https://github.com/NVIDIA-NeMo/Skills/tree/main/recipes}}

    \item We show that scaling \gencluster consistently improves scores, providing a promising path to surpass gold-level performance.

\end{enumerate}

The remainder of this paper is organized as follows. Section 2 provides background on related test-time compute methods and the IOI competition, followed by a detailed breakdown of the GENCLUSTER framework in Section 3. Section 4 presents our experimental results, scaling analyses, and ablation studies, after which we conclude the paper and discuss its current limitations in Sections 5 and 6.

\section{Background}
\label{sec:background}
\subsection{Related Work}
\label{sec:related}

This paper investigates scalable test-time compute (TTC) for large language models (LLMs) applied to coding problems, particularly those with a limited verification budget. Test-time compute refers to allocating additional computational resources during inference to improve output quality. This can be achieved through extended chain-of-thought reasoning~\citep{deepseek_r1, li202512surveyreasoning, Wei2022COT, zhang2025surveytesttimescalinglarge}, generative verifiers for scoring and selection~\citep{fu2025deepthinkconfidence, mahan2024generativerewardmodels, zhang2025generativeverifiersrewardmodeling}, or best-of-N strategies~\citep{chen2025rmr1rewardmodelingreasoning, toshniwal2025genselectgenerativeapproachbestofn, Brown2024LargeLanguageMonkeys,li2022codecontests,Leblond2023AlphaCode2,openai2025competitiveprogramminglargereasoning}. While there exist some limited works that have leveraged large scale test-time compute pipelines to achieve state-of-the-art results on coding problems~\citep{li2022codecontests,Leblond2023AlphaCode2,openai2025competitiveprogramminglargereasoning}, most works have explored smaller compute scales~\citep{chencodet, le2022coderl, zhang2023algosynthesizingalgorithmicprograms} or less challenging benchmarks~\citep{chen2021codex_humaneval, inala2022fault, zhao2025vrank, to2024functional}. These studies have not yet explored their methods at the scale required for competitive programming, where problems are highly complex and only a small fraction of the generated candidates are correct. In the following, we highlight works that specifically explore scalable approaches to competitive programming and coding competitions.

AlphaCode~\citep{li2022codecontests} followed by its successor AlphaCode 2~\citep{Leblond2023AlphaCode2} were the first to demonstrate that large-scale test-time compute can achieve competitive performance on Codeforces contests. AlphaCode achieved results comparable to the top 54\% of participants by generating around 100K solutions per problem and selecting the top 10 via clustering and ranking. AlphaCode 2 improved this further, surpassing 85\% of the participants by scaling up to 1M solutions and refining the selection mechanism. Both used proprietary models and their evaluations were performed on the Codeforces benchmark.

OpenAI then showed gold-medal performance at the International Olympiad in Informatics (IOI) by scaling test-time compute~\citep{openai2025competitiveprogramminglargereasoning}. They fine-tuned a variant of o1 (o1-ioi) for IOI and achieved gold by generating up to 10K solutions per problem and filtering with clustering and heuristic-based selection. They also demonstrated that their o3 model can achieve similar with only 1K solutions per problem and a simpler selection procedure. While they reported achieving gold again at IOI 2025 and a 6th overall ranking, details about the models and selection procedures were not released publicly. Recently, both OpenAI~\citep{thedecoder_openai_ioi2025} and Google DeepMind ~\citep{LinCheng2025GeminiICPC} claimed gold-medal at ICPC 2025 with limited disclosed details. All these works rely on closed models while they provide little to no detail of the test-time compute approach used. 

While these prior systems introduced influential concepts such as large-scale generation and execution-based clustering, they do not address the challenge of selecting strong representatives under strict submission budgets. In this work, we bridge this methodological gap by introducing a novel cluster-level tournament ranking stage and a structured, budget-aware allocation strategy. Furthermore, our work is the first to show that achieving the gold medal at competitions like IOI is feasible with open-weight models and we provide a fully specified, reproducible approach along with systematic scaling analyses to achieve this.

\subsection{IOI Competition and Benchmark}
At the IOI 2025 competition, contestants are presented with six problems, each divided into 2-12 subtasks. Each subtask evaluates specific aspects of the solution under strict time and memory constraints. Among the six problems, five require executable code submissions, while one accepts either executable code or precomputed output. Scoring is determined at the subtask level, with the contestant's final score for a subtask being the maximum achieved across all submissions. Subtask scores are then aggregated to yield a problem score, with each problem worth up to 100 points. Contestants may refine and resubmit their solutions, but each problem allows a maximum of 50 submissions, making submission strategy an essential factor in performance.

Our evaluation dataset is organized at the subtask level rather than the problem level. Each of the six problems is decomposed into its constituent subtasks following the official scoring guidelines. For every subtask, we created a standalone instance by removing all information related to other subtasks, isolating it as an independent evaluation unit. The resulting benchmark, IOI-2025, follows the structure of the publicly available IOI-2024 benchmark~\citep{openr1_ioi_dataset}. For scoring, we execute the solutions using the grader programs provided by the IOI organization with the same memory and time limits.

\section{Proposed Approach}
\label{sec:method}
In this section, we propose \gencluster, a scalable test-time compute approach that improves the performance of LLMs on the IOI competition. \gencluster consists of four stages: (1) parallel generation, (2) behavioral clustering, (3) ranking with tournament, and (4) submission. For each subtask, we first generate a large set of candidate programs in parallel. Given the constraint of 50 submissions per problem, we must efficiently select the most promising candidates from the large pool of solutions. We cluster solutions based on behavioral similarity under several inputs, rank clusters using an LLM-based tournament, and finally employ a round-robin submission policy to maximize scores under submission constraints. The overall pipeline can be seen in Figure~\ref{fig:genclusteroverview}.

\subsection{Parallel Candidates Generation}
We first generate $K$ candidate solutions for each subtask using the prompt in Figure~\ref{fig:solution_generation_prompt}. Since generations are independent, they can be executed in parallel, making our approach scalable. Generations that result in no parsed code or code that does not compile are filtered.

\subsection{Behavioral Clustering}
In this stage, we group candidate solutions based on similar behavior to reduce the number of candidates and make ranking more effective. Inspired by \citet{openai2025competitiveprogramminglargereasoning}, for each subtask, we first instructed the LLM to produce multiple distinct programs that generate randomized test inputs consistent with the subtask's specifications (prompt shown in Figure~\ref{fig:test_data_generator_prompt}). To ensure the correctness of test inputs, we also asked the model to produce multiple independent validators (prompt shown in Figure~\ref{fig:test_data_validator_prompt}). Each validator program checks whether a given input meets the specific constraints of a subtask. Once we have created all test generators and validators, we iterate through the generators. For each one, we produce a test input and evaluate it using all validators. If at least 75\% of the validators approve the input, we keep it as a valid test case. This process is repeated over all the generators until we collect the required number of accepted test cases.

All candidate solutions for a subtask are executed on all the test cases, and then clustered based on their outputs. Solutions that produce exactly the same outputs are grouped into the same cluster. For efficiency, we compute hash values of the outputs to accelerate clustering. If solutions in a cluster produce an empty output for any of the test inputs due to any reason such as runtime errors, that cluster is removed completely.

\subsection{Ranking with Tournament}
Clustering the candidates results in a large number of groups of similarly behaving solutions. We therefore require a ranking mechanism to prioritize the most promising clusters for submission. There has been prior work on selecting the best solution among multiple generated candidates; however, most existing approaches are not well suited to our setting or have not been evaluated under comparable conditions~\citep{liu2025pairjudge, toshniwal2025genselectgenerativeapproachbestofn, Brown2024LargeLanguageMonkeys}. Specifically, (1) they do not scale to the large number of solutions required in our experiments, (2) they are primarily designed and evaluated on tasks such as mathematics, where majority voting can trivially identify the correct answer, or (3) they focus on selection rather than ranking of candidate solutions. In the following section, we present our proposed approach, and report the results of some of the alternative selection strategies in Section \ref{section:rankingablation}.

We propose to hold a tournament between clusters inspired by a partial round-robin tournament. Each cluster plays $G_n$ times with randomly selected clusters. We select the solution with the longest thinking length as the representative solution from each cluster. The representatives of two clusters are then given to the LLM which is prompted to select the better solution in a pairwise comparison (prompt shown in Figure \ref{fig:selection_prompt}). Each play produces a single winner. After the tournament, clusters are then ranked based on the number of wins of their representative candidate. To mitigate recency bias~\citep{shi2024judging} in the model’s judgments, we randomize the presentation order of the two solutions; each solution appears first in approximately half of its matches.

\subsection{Submission Strategy}
We submit up to 50 solutions per problem (the maximum permitted by IOI) using a round-robin strategy. We begin with the final subtask of each problem, which is typically the most difficult. For each subtask, we iterate through its clusters in ranked order, selecting one solution from each cluster and submitting it for scoring. Within each cluster, individual solutions are ranked by their reasoning length, which we use as a proxy for correctness likelihood. We continue submitting one solution at a time from the top-ranked clusters, cycling across clusters in a round-robin fashion and rotating through solutions within each cluster. Once a subtask is solved (i.e., its maximum score is achieved), we skip its remaining clusters and proceed to the next subtask following \cite{openai2025competitiveprogramminglargereasoning}.

\section{Experiments}
\label{sec:experiments}
In our experiments, all solutions, test generators, and validators are generated in C++ since it is the most common language for IOI competition. For each subtask, we produced 100 test generator functions and 100 validators to obtain 100 validated test cases per subtask. We study the impact of varying the number of test cases per subtask in Section \ref{sec:testsetsize}. In all experiments, we used 10 competitions per cluster, $G_n=10$, and studied the impact of this parameter in Section \ref{sec:numofgames}. We used maximum generation lengths of 120K, 120K, 64K, and 120K tokens for models \texttt{gpt-oss-120b}, \texttt{gpt-oss-20b}, \texttt{DeepSeek-R1-0528}, and \texttt{Qwen3-235B-A22B}, respectively. These values correspond to each model’s architectural limit. The effect of the generation length is studied in Section \ref{sec:maxnumoftoken}. To calculate the scores for each solution, we used the graders officially provided by IOI with the exact same time and memory constraints~\citep{ioi_2025_tasks}. 

\begin{figure*}[tb]
    \centering    
    \includegraphics[width=0.8\textwidth]{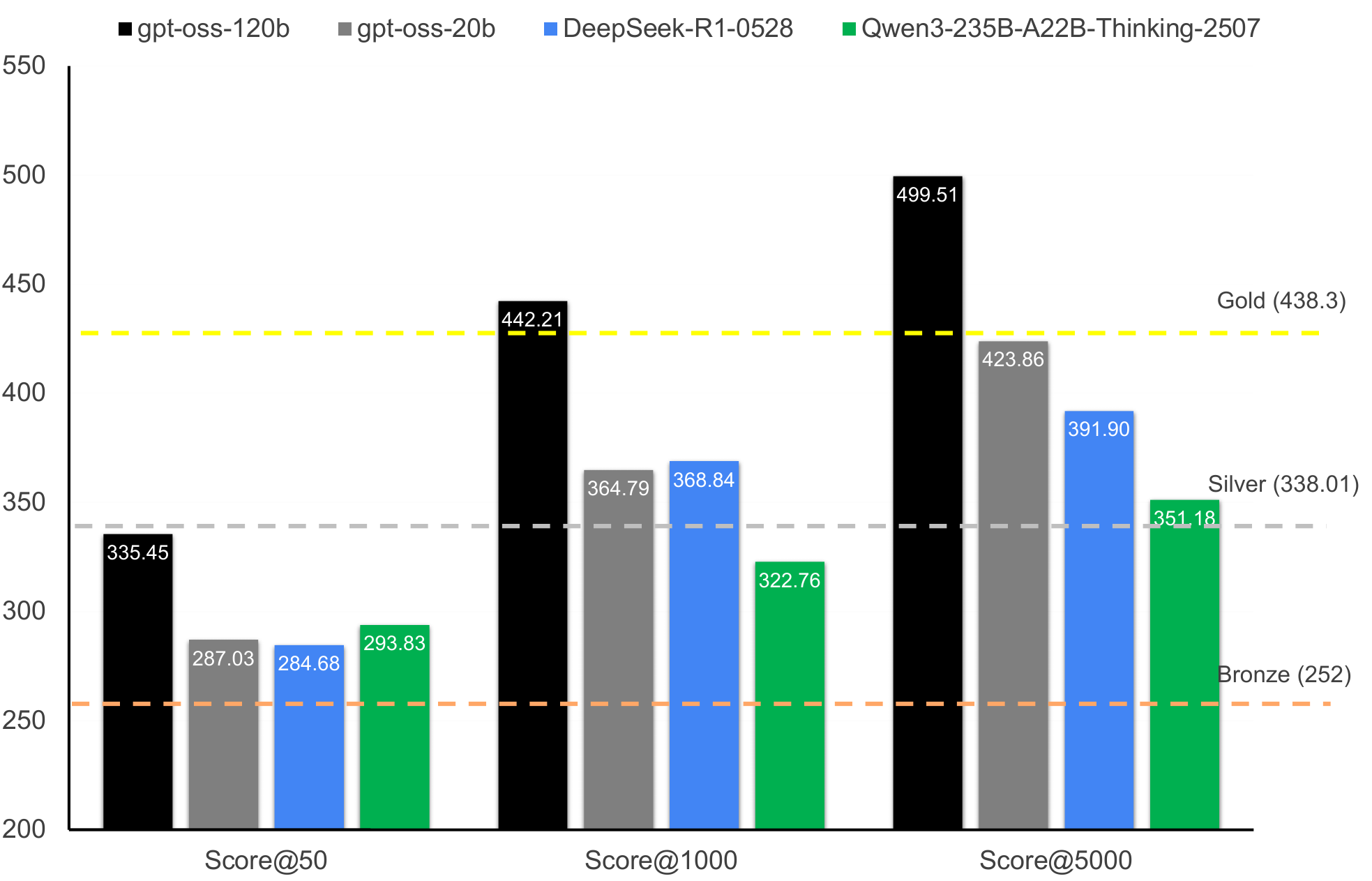}
    \vspace{-2mm}
    \caption{Final scores of different models on IOI 2025 when generating $K$ solutions per subtask, assuming unlimited submissions. Medal thresholds are indicated on the chart, and the maximum achievable score is 600.}
    \label{fig:model_selection}
\end{figure*}

\subsection{Performance Comparison between Models}
\label{sec:model_comparison}
To identify the most competitive open-weight models for the IOI benchmark, we evaluated four top-performing available candidate models: \texttt{gpt-oss-120b}, \texttt{gpt-oss-20b}, \texttt{DeepSeek-R1-0528}, and \texttt{Qwen3-235B-A22B-Thinking}. For each subtask, we generated up to 5000 candidate solutions to estimate the maximum potential performance of each model without submission constraints. The resulting scores on the IOI 2025 benchmark, reported using the \texttt{Score@K} metric, are shown in Figure~\ref{fig:model_selection} for $K \in \{50, 1000, 5000\}$. The results are averaged over 20 runs by randomly sampling from the pool of 5000 generated solutions, except for $K = 5000$, which corresponds to a single run.

As illustrated, \texttt{gpt-oss-120b} achieves the highest score among all models by a significant margin and is the only one with the potential to reach gold medal performance with up to 5000 generations. Moreover, the \texttt{gpt-oss} models exhibit stronger gains as the number of generations increases, suggesting that they scale more effectively with test-time compute. In contrast, while \texttt{Qwen3-235B-A22B-Thinking} outperforms \texttt{gpt-oss-20b} and \texttt{DeepSeek-R1-0528} at smaller generation budgets, its performance scales less favorably.

\begin{figure*}[tbp]
    \centering
    \includegraphics[width=1.0\textwidth]{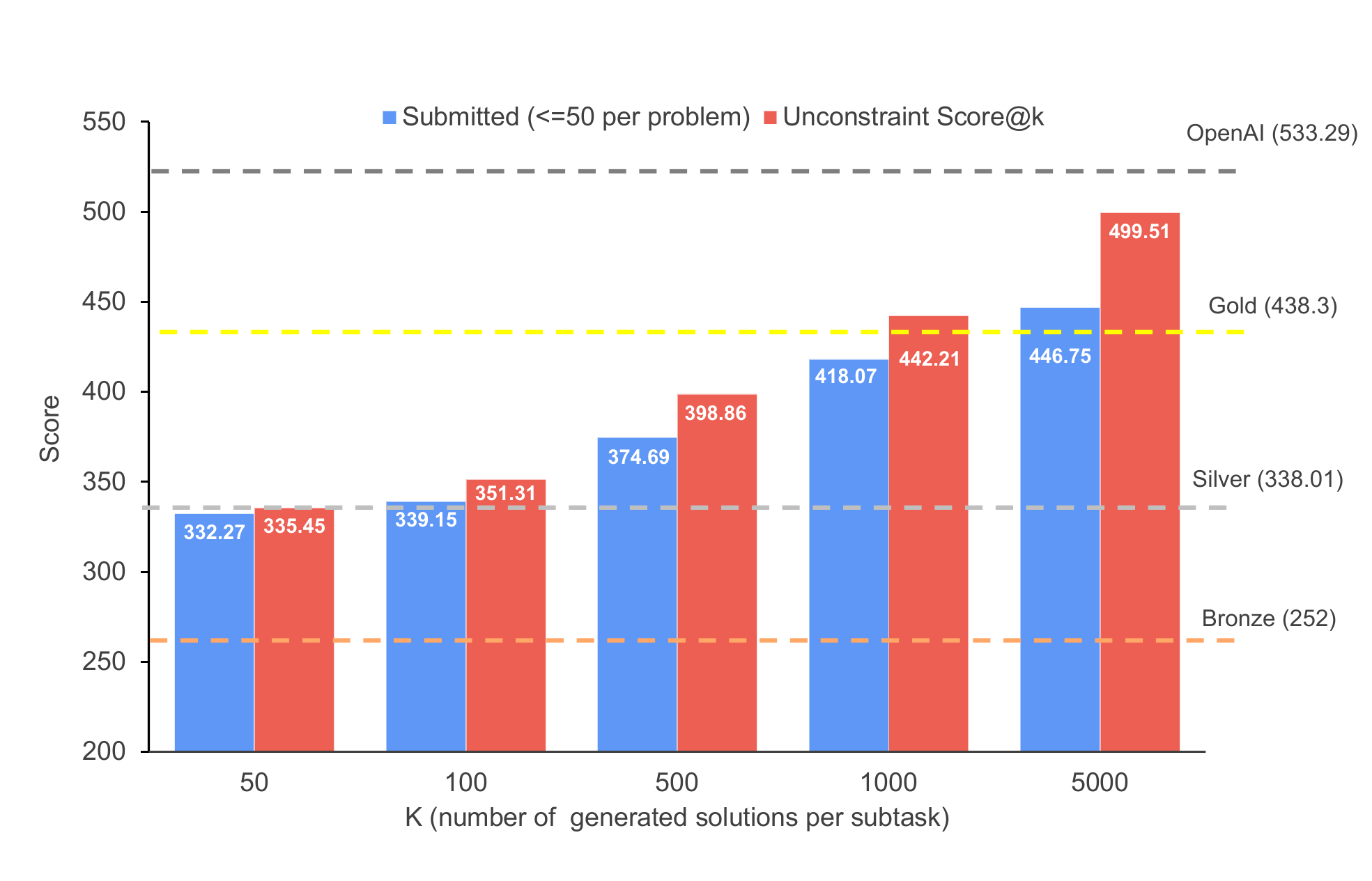}
    \caption{Performance of \texttt{gpt-oss-120b} on IOI 2025 with and without the 50-submission limit, varying generation counts. Results are averaged over five runs (except $K{=}5000$, single run). Medal thresholds are indicated on the chart, and the score reported by OpenAI is shown for comparison.}
    \label{fig:scaling}
\end{figure*}

\subsection{Evaluating \gencluster under Submission Constraints}
We next study the scalability and effectiveness of \gencluster with \texttt{gpt-oss-120b}, the best open-weight model for IOI in our experiments. As shown in Section \ref{sec:model_comparison}, it is the only model that can potentially win a gold medal with up to 5000 generations. In this section, we use \gencluster to follow the submission restrictions of IOI akin to human contestants. Figure~\ref{fig:scaling} shows the scores when using different generation sizes, $K \in \{50, 100, 500, 1000, 5000\}$. We report two outcomes: (1) the \textbf{Submitted Score}, which respects the 50-submission limit by using \gencluster, and (2) the \textbf{Unconstrained Score@K}, which assumes no submission cap and reports the best score among the $K$ solutions. This comparison quantifies the gap between the practical effectiveness of \gencluster and its theoretical upper bound.

As demonstrated, we achieve a gold medal with the \gencluster approach and 5000 generations per subtask while following the constraints of IOI competition. This is the first work to achieve a gold medal with open-weight models at an IOI competition. Although this score remains below the level claimed by OpenAI for IOI 2025, our transparent methodology provides a reproducible baseline for narrowing the gap between the open and closed models on competitive programming tasks. It is also worth noting that OpenAI has not disclosed the compute budget or methodology used to obtain their reported results.

Our results reveal a clear scaling trend: larger candidate pools improve both constrained and unconstrained scores. Even under the 50-submission limit, the submitted score steadily improves from 332.27 for $K = 50$ to 446.75 for $K = 5000$. These results demonstrate the benefits of scaling test-time compute in conjunction with \gencluster. At smaller $K$ values, the gap between the two is modest, but as $K$ increases, the gap gets larger which shows the challenge of selecting the best solutions when we have a large number of candidates. The remaining gap between constrained and unconstrained performance highlights the need for more effective ranking and selection strategies to fully realize the available potential.

\subsection{Comparison with other Test-time Compute Strategies}
\label{section:rankingablation}

\begin{table*}[htbp]
\caption{
Performance comparison of different test-time compute strategies with \gencluster. 
The results show the final scores achieved by each method after submitting up to 50 solutions per problem. 
\textbf{Random} and \textbf{Longest} are non-clustering baselines, while all other methods use clustering. 
“Longest” refers to the solution with the longest reasoning trace.
}
\centering
\rowcolors{1}{white}{gray!9}

\setlength{\tabcolsep}{3.5pt}
\renewcommand{\arraystretch}{1.0}

\begin{tabularx}{\textwidth}{
    >{\RaggedRight\arraybackslash}l
    >{\RaggedRight\arraybackslash}X
    >{\RaggedRight\arraybackslash}X
    >{\RaggedRight\arraybackslash}X
    >{\raggedleft\arraybackslash}r
}
\toprule
\makecell{\textbf{}\\\textbf{Method}} &
\makecell{\textbf{Cluster}\\\textbf{Representative}} &
\makecell{\textbf{Cluster}\\\textbf{Ranking}} &
\makecell{\textbf{Solution}\\\textbf{Selection}} &
\makecell{\textbf{}\\\textbf{Score}} \\
\midrule
Random & -- & -- & Random & 300.10 \\
Longest & -- & -- & Longest & 277.36 \\
Cluster-Size & -- & Size & Longest & 299.87 \\
Cluster-Majority & -- & Majority Voting & Longest & 314.22 \\
\gencluster (Random-Rep) & Random & Number of Wins & Longest & 406.49 \\
\gencluster (Score-Based) & Longest & Average Score & Longest & 441.11 \\
\gencluster & Longest & Number of Wins & Longest & 446.75 \\
\bottomrule
\end{tabularx}
\label{tab:different_test_time_compute}
\end{table*}

We compare \gencluster with several alternative test-time compute selection strategies using 5000 generated solutions per subtask from \texttt{gpt-oss-120b}, evaluated under standard IOI submission constraints. We also include several \gencluster variants as ablations. Results are reported as final scores and summarized in Table~\ref{tab:different_test_time_compute}. The following approaches are considered for comparison:

\begin{itemize}[noitemsep, topsep=0pt]
    \item \textbf{Random}: Selects solutions randomly from all available generated candidates.
    \item \textbf{Longest}: Selects solutions by longest reasoning trace determined by number of tokens.
    \item \textbf{Cluster-Size}: Ranks clusters by size, assuming larger clusters imply higher correctness likelihood.
    \item \textbf{Cluster-Majority}: Ranks clusters based on aggregated majority vote over outputs from generated solutions. Each unique output across the generated test inputs is treated as a vote. For a given test case, we count how many generated solutions produced each particular output. Within each cluster, we then sum the counts of all outputs produced by the generated solutions in that cluster. This prioritizes clusters which collectively produce outputs that are found from the majority of all solutions.
    \item \textbf{\gencluster (Random-Rep)}: A variant of \gencluster where each cluster's representative solution is chosen randomly.
    \item \textbf{\gencluster (ScoreBased)}: A variant of \gencluster where clusters are ranked based on their average score during competitions instead of the number of wins. During comparison, we prompt the model to provide scores for each solution as a number between 0 to 10 (Figure \ref{fig:selection_prompt}).
    \item \textbf{\gencluster}: Our proposed approach which uses the solution with the longest thinking trace as the representative of each cluster and clusters are ranked based on the number of their wins in the tournament.
\end{itemize}

The results clearly show that \gencluster is the most effective method among those evaluated, outperforming all alternatives by a substantial margin. Simple approaches such as selecting solutions with the longest reasoning trace or random selection perform considerably worse. Heuristic-based methods, such as ranking clusters by size (referred to as Cluster-size), are suboptimal and yield scores comparable to random selection, a phenomenon magnified by the difficulty of problems in IOI compared to other benchmarks with easier questions. On such challenging problems where models rarely generate correct solutions, approaches based on majority vote may not be very effective \citep{cobbe2021trainingverifierssolvemath, yang2024qwen25mathtechnicalreportmathematical, wu2025inferencescalinglawsempirical}.

To validate the selection strategy used in \gencluster for choosing cluster representatives, we compare it against a simple baseline that selects a random solution from each cluster (referred to as \textbf{\gencluster (Random-Rep)}). The results show that choosing the solution with the longest reasoning trace performs better than random selection, indicating a positive correlation between reasoning length and correctness within a cluster. Furthermore, ranking clusters by the number of wins provides a slight improvement over using average scores (referred to as \textbf{\gencluster (Score-Based)}).

\subsection{Impact of the Test Set Size on Clustering}
\label{sec:testsetsize}
\begin{figure}[tb]
    \centering    
    \includegraphics[width=1\columnwidth]{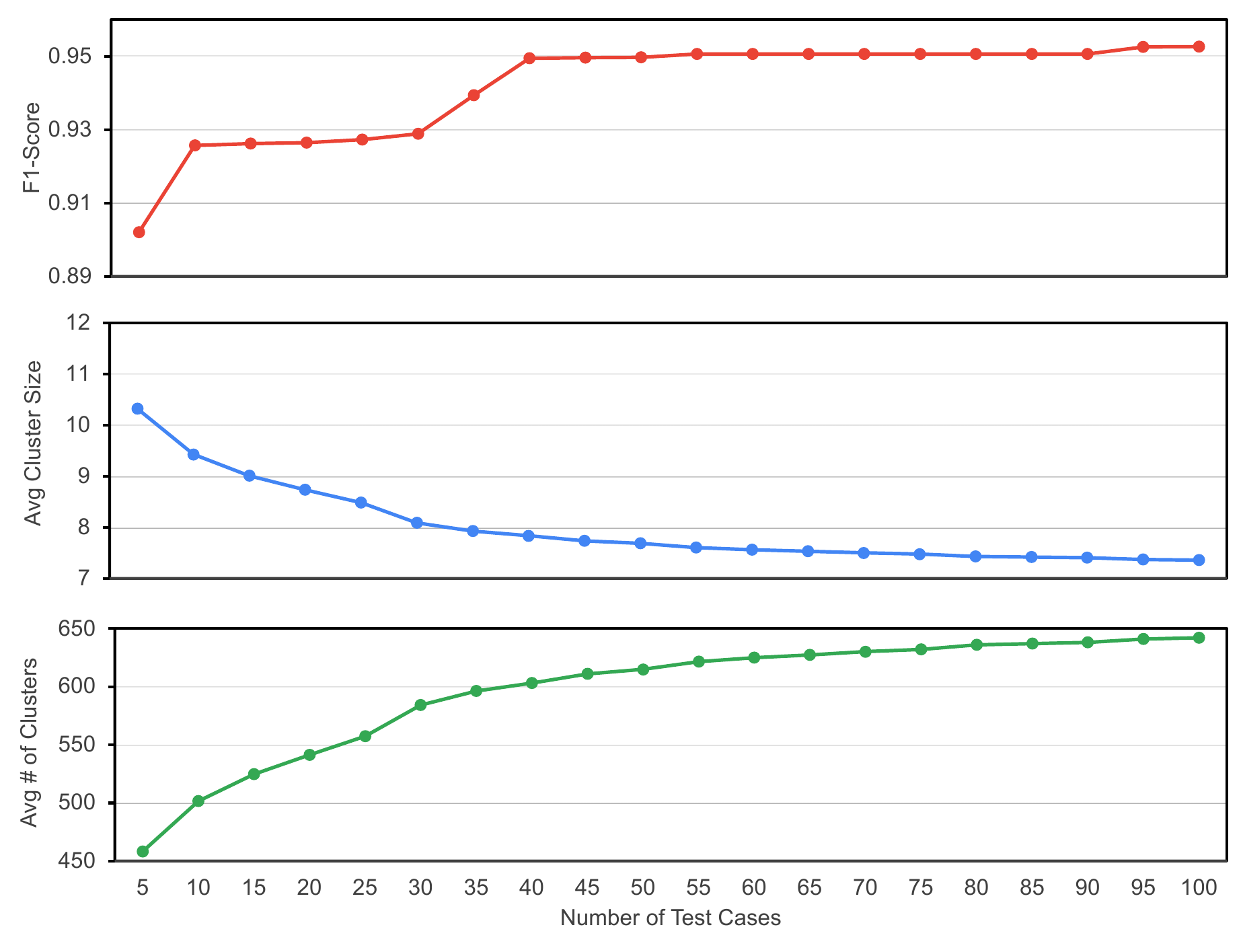}
    \caption{Shown are the cluster purity (F1-score), average cluster size, and average number of clusters for different numbers of test cases.
    Results are reported for $K = 5000$ using \texttt{gpt-oss-120b}.}
    \label{fig:testsets}
\end{figure}

To understand how the number of test cases affects our clustering mechanism, we analyzed the impact of test set size on three key metrics: cluster purity, average cluster size, and the average number of clusters. The results are shown in Figure~\ref{fig:testsets}.

We evaluate cluster purity using the F1-score, which measures how effectively our clustering separates high-quality solutions from others. A ``good'' solution is defined as one that achieves the maximum achieved grade for its subtask, while other solutions are considered as “bad”. The ideal clustering should result in clusters where all the solutions are either all good or all bad. Each solution is assigned a binary label indicating whether it achieves the maximum score. For each cluster, the majority of solutions would determine the prediction of the cluster. The F1-score is then computed by comparing the cluster-level predictions to the ground-truth labels of all solutions, representing the harmonic mean of precision and recall. A higher F1-score indicates purer clusters.

As the number of test cases increases, cluster purity improves, reaching a high F1-score that demonstrates the effectiveness of our clustering approach. However, as more tests enhance cluster purity, they also increase the average number of clusters while reducing the average cluster size. This introduces a key challenge: although additional test cases improve our ability to distinguish correct from incorrect solutions, they also produce a larger number of distinct clusters, making it more difficult to identify the most promising ones under submission constraints.

\subsection{Impact of the Number of Games in Tournament}
\label{sec:numofgames}
\begin{figure}[tb]
    \centering
    \includegraphics[width=1.0\columnwidth]{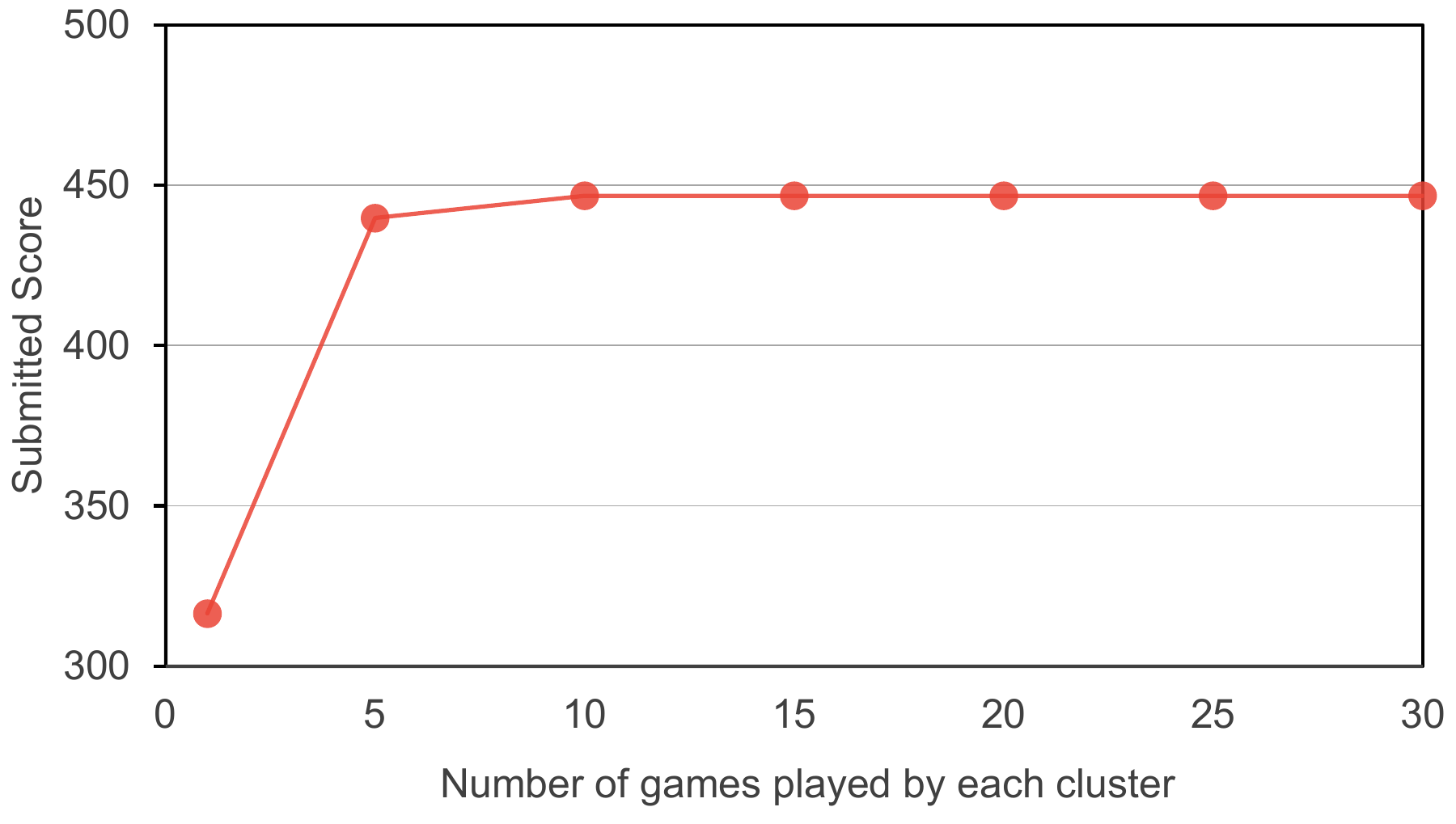}
    \caption{
        Effect of the number of games per cluster on final score.
    }
    \label{fig:games}
\end{figure}

In this experiment, we examine the effect of the parameter $G_n$ (the number of tournament rounds per cluster representative) on the final score under submission constraints. Results for \texttt{gpt-oss-120b} with 5000 generations are shown in Figure~\ref{fig:games}. As shown, increasing the number of games improves the final score but largely saturates after 10 rounds. The gap between one and ten or more games indicates that a single competition is insufficient, and multiple rounds are necessary to obtain reliable judgments and rankings for clusters. Moreover, the observed saturation may reflect inherent limitations of the LLM-as-a-judge paradigm for identifying the best solutions.

\subsection{Evaluating the Quality of the Ranking}
\label{sec:rankingquality}
\begin{figure}[tb]
    \centering
    \includegraphics[width=1.0\columnwidth]{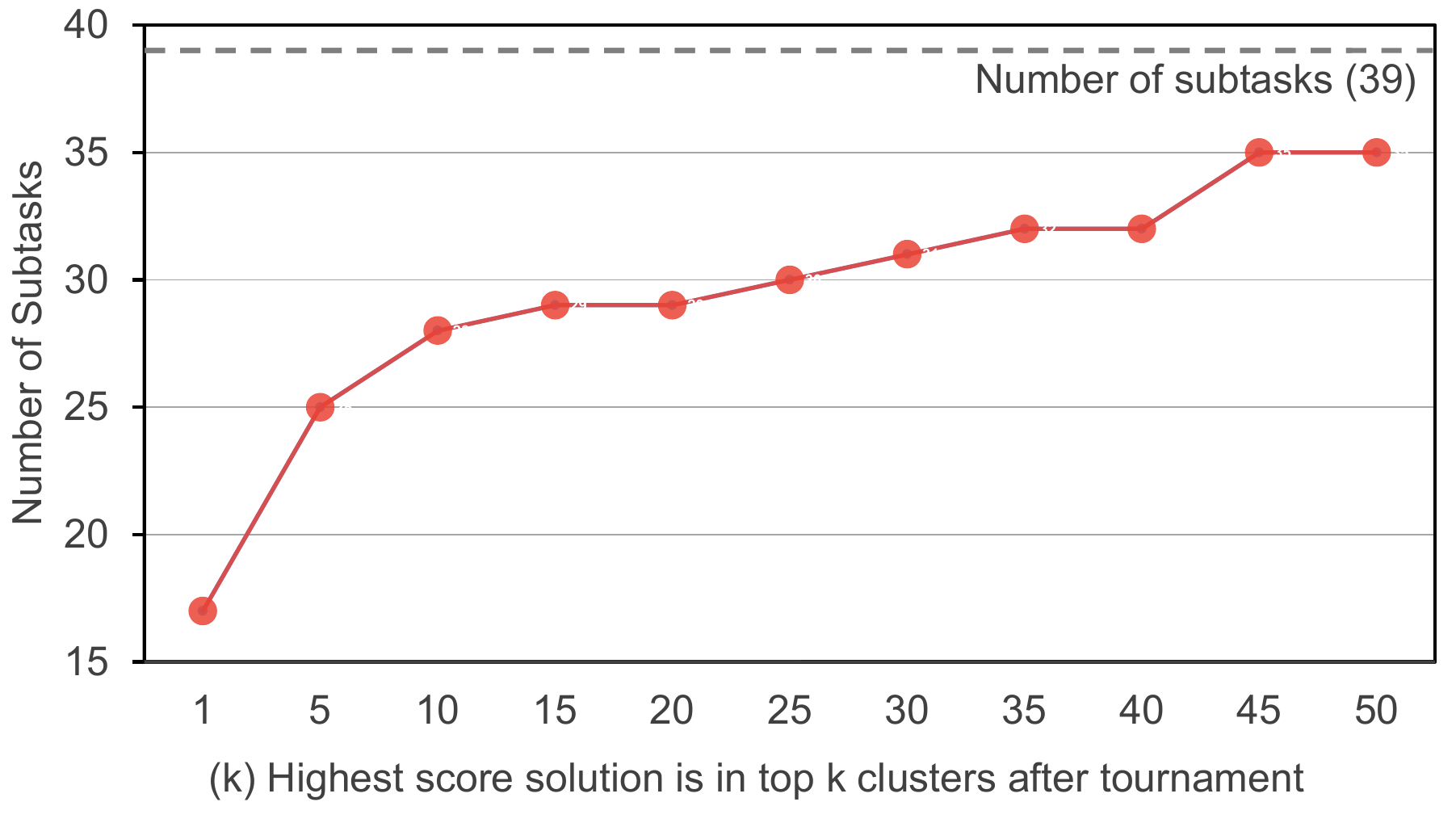}
    \caption{
        Quality of cluster ranking measured by top-$K$ inclusion.
    }
    \label{fig:topk}
\end{figure}

To assess the effectiveness of our cluster ranking method, we evaluated how often the best solution, defined as the one achieving the highest score for a given problem, appeared within the top-K ranked clusters. This metric provides a clear measure of ranking quality. As shown in Figure~\ref{fig:topk}, in 35 out of 39 subtasks, the best solution is included among the top 50 clusters, demonstrating strong performance while still leaving room for further improvement.

\begin{figure}[tb]
    \centering    
    \includegraphics[width=1.0\columnwidth]{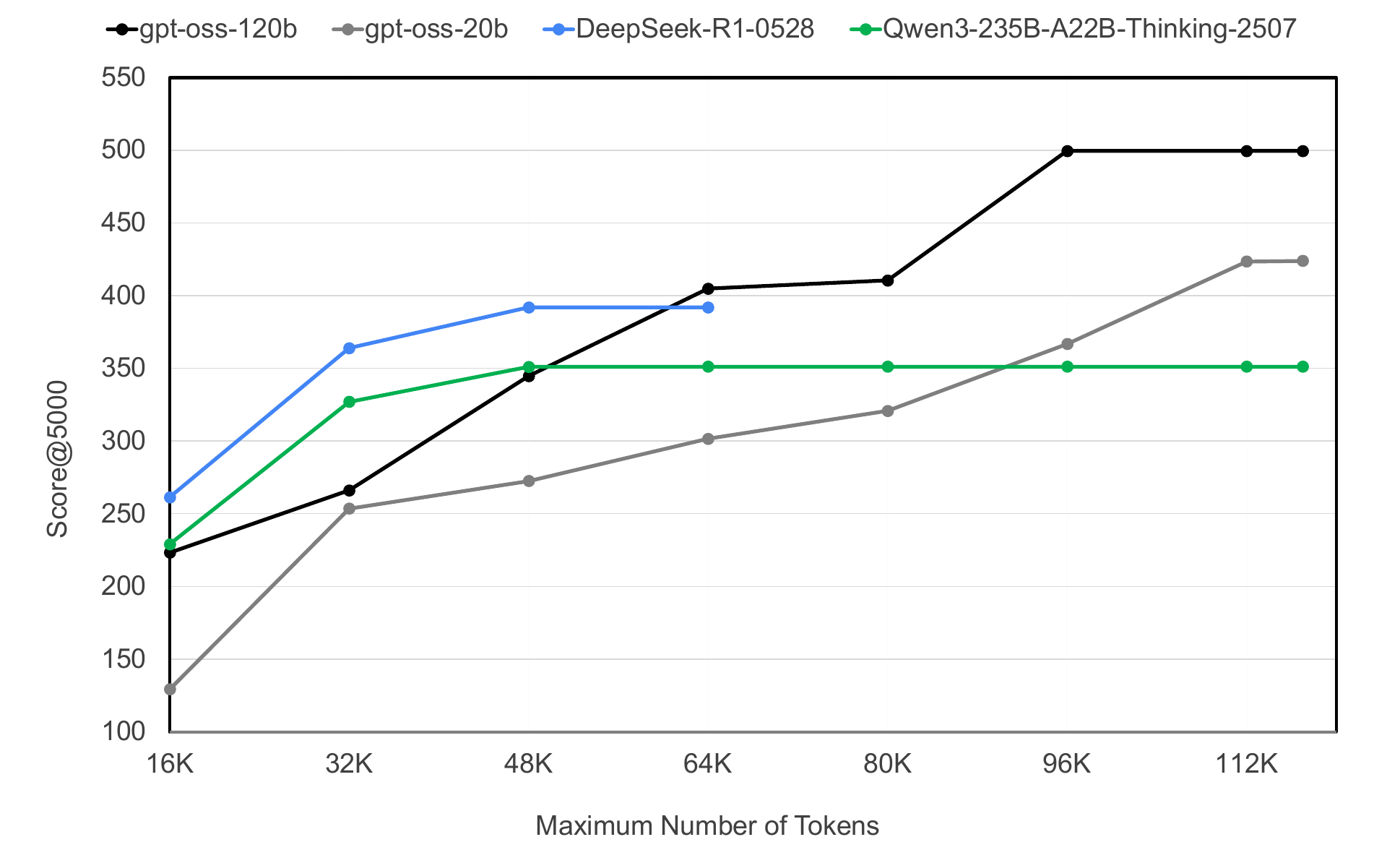}
    \caption{Score@K for different maximum number of tokens in generation with different models.}
    \label{fig:tokens}
\end{figure}
\subsection{Impact of the Maximum Number of Tokens}
\label{sec:maxnumoftoken}
Previous work has shown a correlation between reasoning length and accuracy on challenging problems when reasoning models are used~\citep{OpenAI2024LearningToReason}. In this section, we analyze the relationship between performance and generation length by comparing the \texttt{Score@5000} of different models under varying maximum token limits. Results are presented in Figure~\ref{fig:tokens}, and the rationale behind the chosen generation limits for each model is discussed in Section~\ref{sec:experiments}. The scores of \texttt{gpt-oss-120b}, \texttt{gpt-oss-20b}, and \texttt{DeepSeek-R1-0528} continue to improve up to their respective token limits, whereas \texttt{Qwen3-235B-A22B} saturates after approximately 48K tokens. Notably, the family of gpt-oss models generate longer reasoning traces for difficult problems and achieve stronger performance as a result. While \texttt{DeepSeek-R1-0528} and \texttt{Qwen3-235B-A22B} perform better at shorter reasoning lengths, the gpt-oss models surpass them when provided with greater compute budgets.

\section{Conclusion}
\label{sec:conclusion}
In this work, we present a test-time compute approach that achieves gold-medal performance at the IOI using open-weight models for the first time. Our method generates a large pool of candidate solutions in parallel, followed by a pipeline of behavioral clustering, tournament ranking, and a structured round-robin submission strategy to identify the most promising solutions within the strict IOI submission constraints. Through systematic scaling laws and ablation studies, our experiments show that gpt-oss-120b, when equipped with our test-time compute approach, consistently improves its score on IOI as the number of generated candidates increases. Finally, by providing a fully specified and reproducible pipeline, we establish a strong open-weight baseline that narrows the performance gap with proprietary systems on complex, verifier-limited reasoning tasks.

\section{Limitations}
\label{sec:limitations}
Our results demonstrate that large-scale test-time compute combined with \gencluster can substantially improve open-weight model performance on IOI-style competitive programming. However, several limitations remain.

\paragraph{High Compute and Infrastructure Requirements}
\gencluster relies on generating thousands of candidate solutions per subtask and running an LLM-as-a-judge tournament for ranking. Quantitatively, generating 5,000 candidate solutions for the benchmark requires approximately 7.3 billion tokens, with the ranking tournament consuming an additional 7.3 billion tokens. This substantial requirement for compute and orchestration infrastructure may make the approach impractical for some real-time applications and limits accessibility in low-resource environments.

\paragraph{Dependence on Synthetic Test Generation and Incomplete Behavioral Coverage}
Behavioral clustering relies on model-generated test cases and validators. While validator agreement reduces obvious invalid tests, the resulting test sets may still have blind spots and fail to distinguish solutions that differ only on rare corner cases. Consequently, some incorrect solutions may cluster with correct ones (or vice versa), and the diversity of tests required to separate behaviors grows quickly with problem complexity.

\paragraph{Imperfect Ranking and LLM-as-a-judge}
Our tournament ranking uses an LLM to compare solution pairs based on code and reasoning traces. Such judgments can be noisy and influenced by superficial features (e.g., code style, length, or explanation quality) rather than actual correctness. Although randomizing presentation order reduces position bias, misrankings remain possible and contribute to the observed gap between unconstrained Score@K and submitted scores under the 50-submission limit.

\paragraph{Imperfect Correlation between Reasoning Length and Correctness}
We use reasoning length to select cluster representatives and order candidates within clusters. While this heuristic improves over random selection in our experiments, longer traces can also reflect confusion, verbosity, or unproductive exploration. This makes it unreliable in some cases, potentially over-prioritizing verbose but incorrect solutions.

Overall, \gencluster provides a scalable path to improve competitive programming performance with open-weight models, but its resource demands, reliance on synthetic tests, and ranking noise indicate clear opportunities for improving efficiency, robustness, and accessibility.
\bibliography{GenCluster}

\newpage
\appendix
\onecolumn
\section{Prompts}
\label{appendix:prompts}

\begin{figure}[htb]
\centering

\begin{tcolorbox}[title={Solution Generation Prompt}, colback=red!0, left=2pt,right=2pt,top=2pt,bottom=2pt]

{ 
You are an expert competitive programmer. You will be given a problem statement, test case constraints and example test inputs and outputs. Please reason step by step about the solution, then provide a complete implementation in C++17. You should correctly implement the routine(s) described in Implementation Details, without reading or writing anything directly from stdin or to stdout, as input and output are passed through the implemented routines. Assume your code will be run on the OFFICIAL grader, and do not add a main function, a sample grader, or any other functionality unless it has been explicitly requested.
\vspace{0.2cm}

Put your final solution within a single code block:

\verb|```|cpp\\
// your code here\\
\verb|```|
\vspace{0.2cm}

\{question\}

}
\end{tcolorbox}
  \caption{The prompt used for generating the solutions}
  \label{fig:solution_generation_prompt}
  \vspace{-0.2in}
\end{figure}

\begin{figure}[htb]
\centering

\begin{tcolorbox}[title={Test Data Generator Prompt}, colback=red!0, left=2pt,right=2pt,top=2pt,bottom=2pt]

{ 
You are an expert competitive programmer. You will be given a problem statement, its constraints, and example test cases.  Your task is to write a **test case generator** in C++17 that produces valid inputs for the problem, following the constraints and reflecting the variety suggested by the examples.  
\vspace{0.2cm}

First, **reason step by step** about how to design the generator.  
\vspace{0.2cm}

Then, provide the **final complete implementation** inside a single code block:

\verb|```|cpp\\
// your code here\\
\verb|```|

\vspace{0.2cm}

\#\#\# Question:
\vspace{0.2cm}

\{problem\}

\vspace{0.2cm}
\#\#\# Answer: (use the provided format with backticks)
}
\end{tcolorbox}

  \caption{The prompt used for generating the test data generators}
  \label{fig:test_data_generator_prompt}
  \vspace{-0.2in}
\end{figure}

\begin{figure}[htb]
\centering

\begin{tcolorbox}[title={Test Data Validator Prompt}, colback=red!0, left=2pt,right=2pt,top=2pt,bottom=2pt]

{ 
You are an expert competitive programmer. You will be given a problem statement, its input format, constraints, and examples. Your task is to write an **input validator** in C++17 that reads from stdin and checks that the input fully matches the spec and all constraints.  
\vspace{0.2cm}

First, reason step by step about what must be validated.  
\vspace{0.2cm}

Then, provide the **final complete implementation** inside a single code block: 

\verb|```|cpp\\
// your code here\\
\verb|```|

\vspace{0.2cm}

The program should print ``passed'' if the input is valid, otherwise ``failed'' and a short error to stderr. It must also ensure no extra tokens remain.  
\vspace{0.2cm}

\#\#\# Question:
\vspace{0.2cm}

\{problem\}
\vspace{0.2cm}

\#\#\# Answer: (use the provided format with backticks)
}
\end{tcolorbox}
  \caption{The prompt used for generating the test data validators}
  \label{fig:test_data_validator_prompt}
\end{figure}

\begin{figure}[htb]
  \centering

\begin{tcolorbox}[title={Solution Selection Prompt}, colback=red!0, left=2pt,right=2pt,top=2pt,bottom=2pt]

{ 
You are an expert competitive programmer. You will be given a problem statement, its constraints, and two solutions.  Your task is to evaluate each solution's correctness based on the problem statement and its constraints and select the best solution.  

First, **reason step by step** about each solution. \\
\vspace{0.2cm}

\#\#\# Question: \\
\{problem\}
\vspace{0.2cm}

\#\#\# Solution A\\
\{code\_A\}
\vspace{0.2cm}

\#\#\# Solution B\\
\{code\_B\}
\vspace{0.2cm}

Finish your reasoning with exactly three lines, nothing else:\\
Score A: \textless 0-10\textgreater\\
Score B: \textless 0-10\textgreater\\
Judgment: [A]  or  Judgment: [B]
}
\end{tcolorbox}

  \caption{The prompt used for comparing the solutions in the tournament}
  \label{fig:selection_prompt}
  \vspace{-0.2in}
\end{figure}

\clearpage
\section{Best Scores on IOI-2025}
\label{appendix:scores}

\sisetup{
  table-number-alignment = center,
  group-minimum-digits = 3,
  round-mode = places,
  round-precision = 2
}
\renewcommand{\arraystretch}{1.05}

\begin{table}[h]
\caption{Scores of the top gold and silver medalists at IOI 2025, compared with the results achieved by \gencluster and OpenAI~\citep{ioi_2025_results}. Per-problem scores and overall rankings are shown.}
\centering
\setlength{\tabcolsep}{4pt}
{\small
\begin{adjustbox}{max width=\textwidth}
\begin{tabular}{
  r    
  l    
  S    
  S    
  S    
  S    
  S    
  S    
  S    
  l    
}
\toprule
\textbf{Rank} & \textbf{Name} &
\textbf{S} & \textbf{T} & \textbf{W} & \textbf{F} & \textbf{M} & \textbf{O} &
\textbf{Total} & \textbf{Medal} \\
\midrule
1  & Hengxi Liu                        & 100 & 100.00 & 100 & 100 & 91.23 & 100 & 591.23 & Gold \\
6  & \textbf{OpenAI}                           & 100 &  75.29 & 93  &  100 & 65   &  100 & 533.29 & Gold \\
26  & \textbf{\gencluster}                        & 100 &  87.66 & 72  &  82 & 22.09 &  83 & 446.75 & Gold \\
29 & Maxim Tsoy                             & 100 &  44.83 & 100 &  82 & 72.89 &  37 & 436.72 & Silver \\
\bottomrule
\end{tabular}
\end{adjustbox}
}
\end{table}

\end{document}